\pdfoutput=1
\documentclass[sigconf]{acmart} 
\long\def\comment#1{}
\usepackage[ruled]{algorithm2e} 

\usepackage{algorithmic}
\usepackage{afterpage}
\usepackage{multirow}
\usepackage{refcount}
\usepackage{enumitem}
\usepackage{subcaption}
\usepackage{footnote}

\copyrightyear{2020}
\acmYear{2020}
\setcopyright{acmcopyright}
\acmConference[CIKM '20] {The 29th ACM International Conference on Information and Knowledge Management}{October 19--23, 2020}{Virtual Event, Ireland}
\acmBooktitle{The 29th ACM International Conference on Information and Knowledge Management (CIKM '20), October 19--23, 2020, Virtual Event, Ireland}
\acmPrice{15.00}
\acmDOI{10.1145/3340531.3412039}
\acmISBN{978-1-4503-6859-9/20/10}

\begin{document}
\fancyhead{} 

\title{Meta-Learning for Neural Relation Classification with Distant Supervision}

\author{Zhenzhen Li}
\affiliation{%
  \institution{National University of Defense Technology}
  \streetaddress{No.109, Deya Road}
  \city{Changsha}
  \country{China}
  \postcode{410073}
}
\email{lizhenzhen14@nudt.edu.cn}

\author{Jian-Yun Nie}
\affiliation{%
  \institution{University of Montreal}
  \city{Montreal}
  \country{Canada}}
\email{nie@iro.umontreal.ca}

\author{Benyou Wang}
\affiliation{%
 \institution{University of Padua}
 \city{Padova}
 \country{Italy}}
 \email{wang@dei.unipd.it}

\author{Pan Du}
\affiliation{
  \institution{University of Montreal}
  \city{Montreal}
  \country{Canada}}

\email{pandu@iro.umontreal.ca}

\author{Yuhan Zhang}
\affiliation{
  \institution{Beihang University}
  \city{Beijing}
  \country{China}
  }
\email{yh_zhang@buaa.edu.cn}

\author{Lixin Zou}
\affiliation{
  \institution{Baidu Inc.}
  \city{Beijing}
  \country{China}
  }
\email{zoulixin15@gmail.com}

\author{Dongsheng Li}
\affiliation{\institution{National University of Defense Technology}
\city{Changsha}
\country{China}
}
\email{dsli@nudt.edu.cn}

\begin{abstract}

Distant supervision provides a means to create a large number of weakly labeled data at low cost for relation classification. However, the resulting labeled instances are very noisy, containing data with wrong labels. Many approaches have been proposed to select a subset of reliable instances for neural model training, but they still suffer from noisy labeling problem or underutilization of the weakly-labeled data.
To better select more reliable training instances,
we introduce a small amount of manually labeled data as reference to guide the selection process.
In this paper, we propose a meta-learning based approach, 
which learns to reweight noisy training data under the guidance of reference data.
As the clean reference data is usually very small, we propose to augment it by dynamically distilling the most reliable elite instances from the noisy data. Experiments on several datasets demonstrate that the reference data can effectively guide the selection of training data, and our augmented approach consistently improves the performance of relation classification comparing to the existing state-of-the-art methods.

\end{abstract}



\maketitle
\section{Introduction}

\begin{figure}
    \centering
    \includegraphics[width=0.99\columnwidth]{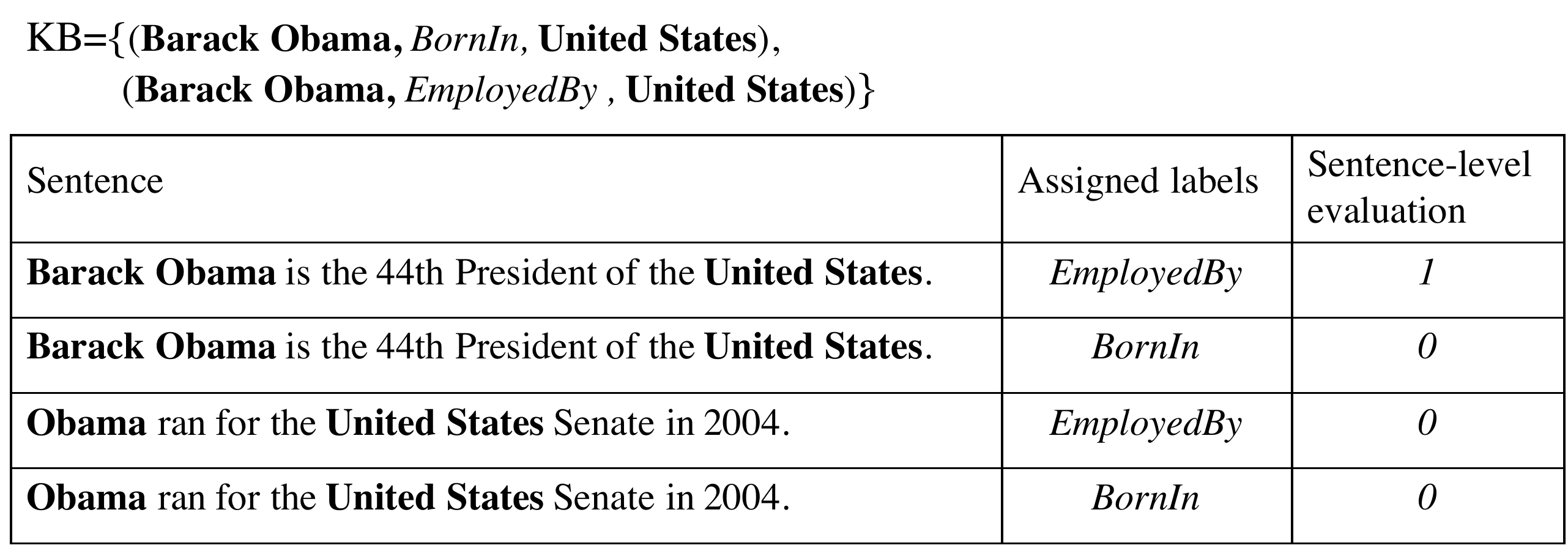}
    \caption{Examples of sentence-level evaluation for DS labeled data, which evaluates each instance individually. ``1'' and ``0'' denote the confidence score about the labelling quality.  
    }
    \label{fig:sentence_level_evaluation_task}
\end{figure}

Relation classification~(RC) aims to categorize the semantic relation between two entities in a sentence into a set of predefined relation types. The task is  useful in various downstream applications, such as information retrieval~\cite{wu2010open}, question answering~\cite{yao2014information} and knowledge graph completion~\cite{zhang2017position}.  Most recent methods tackle this problem by training neural classification models in a supervised way ~\cite{zeng2014relation,zhou2016attention,wang2016relation,zhang2017position}, demanding a large amount of labeled training data which is difficult to acquire in practice. 
To alleviate the problem, distant supervision~(DS) was proposed to generate abundant weakly labeled data~\cite{wu2007autonomously,mintz2009distant} by aligning facts in a knowledge base~(KB) and unlabeled sentences: 
If a fact in the KB states a relation between two entities, then every sentence containing  the same pair of entities is labeled with that relation. 
This process inevitably introduces many wrong labels when the sentences do not express the labeled relation. 
As shown in Fig.~\ref{fig:sentence_level_evaluation_task}, if \emph{BornIn} and \emph{EmployedBy} are the relations between the entity pair~(Barack Obama, United States) in the KB, then every sentence mentioning the same entities will be assigned both relation labels. This process generates many wrong labels (0 in Fig.~\ref{fig:sentence_level_evaluation_task}). 
Training a classifier with such highly noisy labeled data will strongly limit the effectiveness of the classifier.

\begin{figure}
    \centering
    \includegraphics[width=0.98\columnwidth]{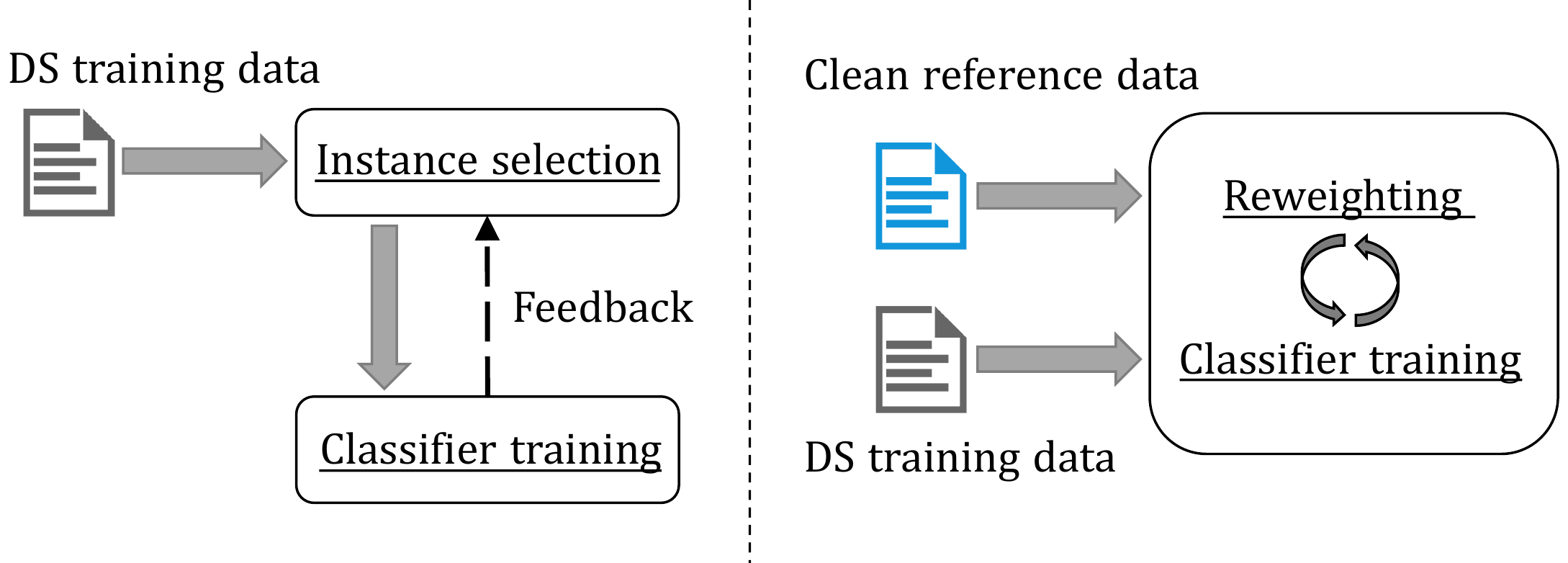}
    \caption{
    Bootstrapping process from noisy data (left) vs. our meta-learning  approach guided by reference data (right).
    }
    \label{fig:sentence_level_noise_reduction}
\end{figure}

To alleviate the impact of wrong labels on the classifier, many existing studies~\cite{zeng2014relation,lin2016neural,ji2017distant,liu2017soft,WangXQ18dsgan,li2019gan,WangXQ18robust,feng2018reinforcement,jia2019arnor} try to select a subset of reliable instances for training. Some  approaches~\cite{zeng2014relation,lin2016neural,ji2017distant,liu2017soft} adopt multi-instance learning, which relaxes the relation label for each sentence to a  
bag of sentences mentioning the same entity pair and assumes that at least one sentence expresses the relation. 
These approaches are often used in relation extraction, 
i.e. to identify possible relations between a pair of entities. However, for our task of sentence-level relation classification (i.e. to judge if a given sentence expresses a specific relation), they turn out to be ineffective~\cite{feng2018reinforcement,jia2019arnor}, because we have to judge on each of the instances rather than a bag.
From this perspective, sentence-level relation classification is fine-grained and thus more difficult than relation extraction. 
Some recent approaches~\cite{feng2018reinforcement,WangXQ18robust,WangXQ18dsgan,han2018denoising,zeng2018large,jia2019arnor} perform sentence-level evaluation. They evaluate the labeling quality of each instance and select those deemed reliable for model training. 
These approaches leverage the large amount of distantly labeled instances to emerge strong supervision signals during the iterative training process. 
As illustrated in the left part of Fig.~\ref{fig:sentence_level_noise_reduction}, they usually resort to reinforcement learning or adversarial learning to train an instance selection model by receiving feedback from the classifier or a manually
crafted reward function. The basic assumption is that most training instances are true, and thus the iterative learning process will help select confident instances whose predicted label by the learned classifier is consistent with its DS-generated label.
We put these approaches in the category of \emph{bootstrapping} approaches, i.e. an instance selection model emerges itself from the noisy data gradually.
Even though the bootstrapping process can learn a good instance selection model, it may be easily trapped by some common wrong instances.  In addition, the instance selector may fail to select some true instances that are different from the frequent instances, 
thus underutilize the useful training data. Existing approaches either retain too many noisy training instances or miss a lot of effective training data, limiting the capability of the resulting classifier. 

The above problem stems from the fact that the selector is trained without further supervision signals. If we have some manually labeled instances that show good examples to express a specific relation in natural language, then the selection model can be guided by them. So, our idea in this paper is to use a small set of manually annotated samples as ``reference data'' to guide the instance selection. This corresponds to a new setting of DS for relation classification, where a small set of reference data is available in addition to a large amount of noisy DS data. This setting is realistic - in practice, it is not difficult to label a small set of data if this turns out to be beneficial, which will be shown in our experiments.

In this paper, we focus on effective utilization of the small set of reference data for relation classification using DS. 
We propose a meta-learning framework, in which the reference data is used to guide instance 
weighting at the meta-learning level. Then the 
weighted instances are used to train a classifier.
We use a specific meta-learning method similar to the online reweighting algorithm~\cite{RenZYU18,shu2019meta}, 
which performs a meta gradient descent step to adjust the  instance weights in order to minimize the classification loss on the clean reference data. 
This method is used because it can more effectively leverage clean reference data than
existing reinforcement learning or adversarial learning approaches~\cite{feng2018reinforcement,WangXQ18dsgan,WangXQ18robust,zeng2018large,han2018denoising}.

As we assume that only a small set of clean data is available, we may have the mode collapse issues: it may over-boost the noisy samples that agree with the reference samples, leading the classifier to collapse into the existing mode and thus a poor generalization capability. 
To relieve this issue, we propose to augment the clean reference data by a set of noisy instances deemed confident. We call them ``elite instances". 
At each training iteration, the elite instances are used together with the clean data to guide a \textit{reweighting} process to assign weights to noisy instances, based on which training instances are selected. The process can combines the strengths of bootstrapping from noisy data and guidance by reference data.
The whole training process iterates between reweighting and classifier training as shown in the right side of Fig.~\ref{fig:sentence_level_noise_reduction}. 

The contributions in this work include: 
\begin{itemize}[leftmargin=*]
  \item We propose an effective way to select noisy data for training, guided by a small set of reference data.
  \item We adopt the meta-learning mechanism for DS relation classification for the first time, and show its high effectiveness compared to other alternative methods.
  \item Our approach combines the strengths of supervision and bootstrapping to create a larger set of reference data for meta-learning.
 \end{itemize}

\section{METHODOLOGY}
\label{sec:problem}

\textbf{Problem Definition} The relation classification problem can be formulated as follows: Let  $\mathbb{D}_\textrm{train}$ be a set of sentence-relation pairs $\{(x_i,r_i)\}_{i=1}^{N}$, where $r_i$ is a relation label created by distant supervision. 
The goal is to train a relation classification model $\Phi$ parameterized by $\theta$ that can predict the relation of a new sentence $x_j$ from test data $\mathbb{D}_\textrm{test}$, i.e., compute the probability $P_{\Phi}(r_j|x_j)$.
Formally, the task of learning under distant supervision is to minimize the empirical risk on training data as follows:  

\begin{equation}
\label{eq:weighted_loss}
\begin{aligned}
\theta ^* 
&= \underset{\theta}{\mathrm{argmin}} \left(  \mathbb{E}_{(x,r) \sim p_{\mathbb{D_\textrm{train}}}}( \ell(\Phi_{\theta}(x),r))\right)\\
&= \underset{\theta}{\mathrm{argmin}} \left(
\sum_{(x,r)} p_{\mathbb{D_\textrm{train} }}(x,r) \ell(\Phi_{\theta}(x),r)\right)\\
&\approx \underset{\theta}{\mathrm{argmin}}  \left(\sum_{i=1}^Nw_i \ell(\Phi_{\theta}(x_i),r_i)\right). \\
\end{aligned}
\end{equation}
In the last step, 
the training data is used with their weights  $\{w_i\}_{i=1}^{N}$ representing the labeling quality. 

The selection of good instances is a key problem in DS learning. Different from the previous approaches~\cite{feng2018reinforcement,WangXQ18robust,WangXQ18dsgan,zeng2018large,han2018denoising}, we assume that we have a small set of clean reference data $\mathbb{D}_\textrm{ref}$ that can be used to guide the selection or weighting process.

\subsection{Overview of the Proposed Approach}

We propose to adopt a  weighting schema of training samples to dynamically reweight 
noisy instances in $\mathbb{D}_\textrm{train}$ under the guidance of  $\mathbb{D}_\textrm{ref}$.
We achieve this goal by proposing an approach based on meta-learning, which aims at minimizing the meta-objective:

\begin{equation}
\label{Eq:overview_meta_obj}
    w ^* = \underset{w}{\mathrm{argmin}} ~
    \mathrm{MetaObjective}(
    \Phi_\theta,\mathbb{D_\textrm{ref}} )\\
\end{equation}

The principle of our meta-learning is as follows: we want the classifier trained with the 
weighted instances to minimize the loss on the reference data. Note that $\Phi_\theta$ is a function of $w$ since $w$ affects the optimization of $\theta$ as shown in Eq.~\ref{eq:weighted_loss}. We will provide details about this instance reweighting~(meta-reweighting) process later.

As we mentioned earlier, we assume that we only have a small set of reference data. A typical consequence is mode collapse, i.e., the selected training instances~(assigned high weights) become similar to the reference data, leading to a model with narrow coverage.
To cope with this challenge, we increase the coverage of the reference data by augmenting it with some highly reliable instances from the noisy set, which we call ``elite data'' or ``elite instances''.
The elite data is expected to  represent some strong patterns of a relation among the noisy data, which may not be covered by the clean reference data.

\begin{figure}[pt]
    \centering
    \includegraphics[width=0.8\columnwidth]{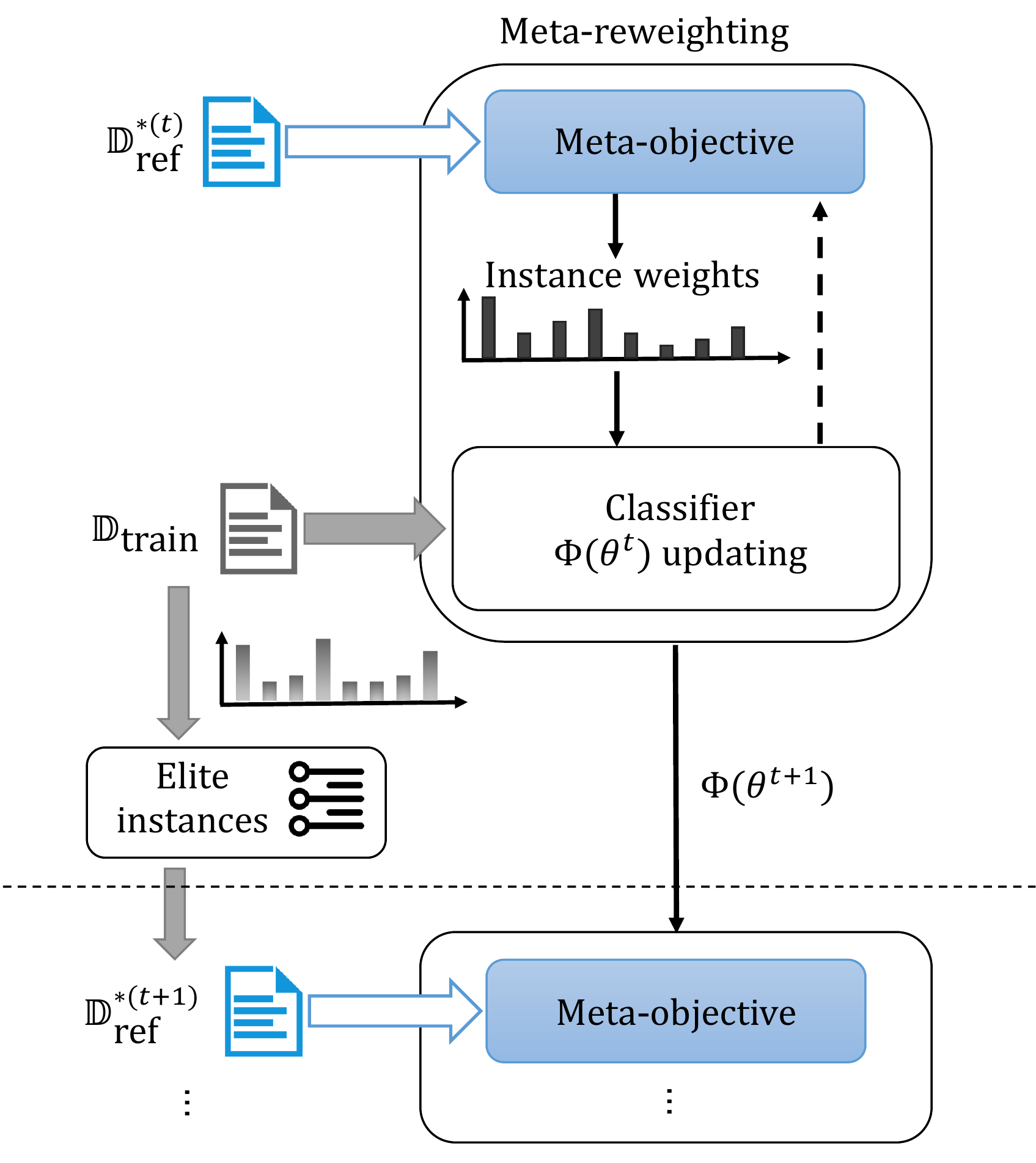}
    \caption{Framework overview. 
    At each training epoch $t$, we learn to reweight noisy $\mathbb{D}_\textrm{train}$ according to a meta-objective with respect to the augmented reference data. 
    The classifier $\Phi(\theta^{t})$ is updated with the weighted instances. 
    Afterwards, we select a set of elite instances to augment the clean reference data to $\mathbb{D}_\textrm{ref}^{\ast(t+1)}$ for the next epoch.
    }
    \label{fig:overview}
\end{figure}

The framework of our proposed approach is shown in the Fig.~\ref{fig:overview}, consisting of iterations of \textbf{robust classifier training} based on meta-reweighting and \textbf{elite instance selection} to augment reference data~(to extend reliable supervision signals).
Details are given in the following subsections.

\subsection{Robust Classifier Training by Meta Learning}
\label{sec:metalearning}
\begin{figure}[t]
\centering
\includegraphics[width=0.8\linewidth]{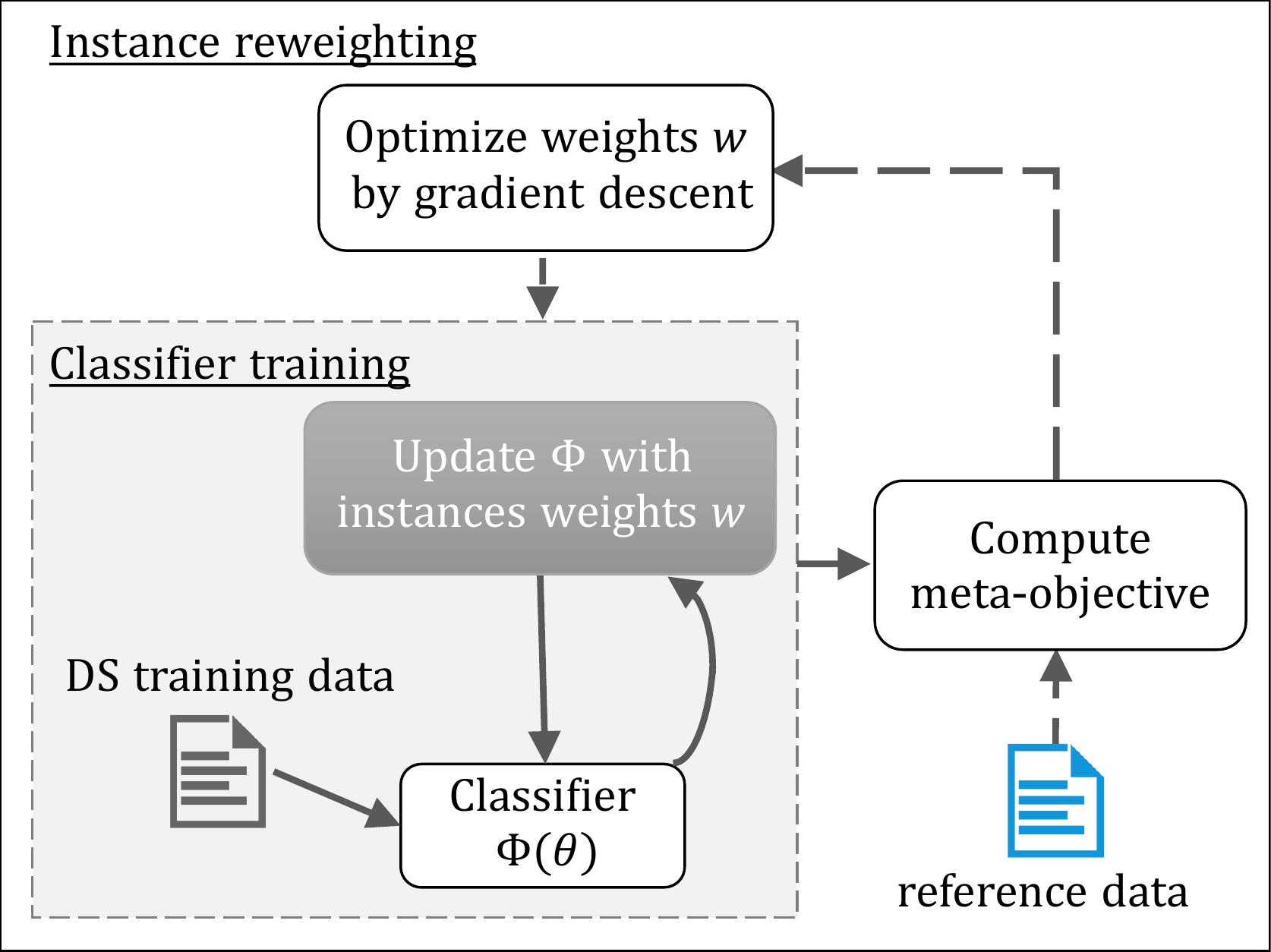}
\caption{Schematic diagram of meta-learning algorithm for robust classifier training, which iterates between instance reweighting and classifier training. The instance reweighting process contains extra classifier training process.  
} 
\label{fig:meta-reweighting}
\end{figure}

Generally, the goal of machine learning is  to find a model $\Phi$ with the best parameter $\theta$ to minimize the empirical risk as in Eq.~\ref{eq:weighted_loss}. In the DS setting, this means to weight the training instances~($w$) to train a better model $\Phi$. We propose to leverage the small set  $\mathbb{D}_\textrm{ref}$ for this purpose using meta-learning based sample reweighting algorithm~\cite{RenZYU18,shu2019meta}. 
As shown in the Fig.~\ref{fig:meta-reweighting}, our meta-learning algorithm involves two optimization process: the outer loop for instance reweighting and the inner loop for classifier training. As the outer loop is optimized for better inner loop, the classifier training is embedded in the instance reweighting process. 
That is, at each iteration,
we start with the classifier training with weighted training data, and then optimize 
$w$ based on the updated classifier with respect to the meta-objective (on reference data). Once $w$ is optimized, we use it to update the model $\Phi$ in the classifier training. The whole training process iterates between the instance reweighting and classifier training. 

To increase the training efficiency, we adopt the online reweighting strategy~\cite{RenZYU18}, which dynamically learns the instances weights for a mini-batch of training data by a single optimization step. 
In the \textbf{instance reweighting} phase, given a mini batch of training instances $\mathbb{D}_\textrm{mtrain} \subset \mathbb{D}_\textrm{train}$ and weight vector $w$ (which may be initialized by perturbing~\cite{koh2017understanding}), we define the weighted training loss as follows:

\begin{equation}
\label{eq:weighted_training_loss}
\mathcal{L}_\textrm{mtrain}\left( \theta(w) \right) = \sum_{(x_i,r_i) \in \mathbb{D}_\textrm{mtrain}}w_i \ell(\Phi_{\theta}(x_i),r_i).
\end{equation}
We update $\theta(w)$ to  a temporary version $\hat{\theta}(w)$ as Eq. \ref{eq:temp_theta},  which is only used to optimize instance weights:
\begin{equation}
\label{eq:temp_theta}
\hat{\theta}(w) = \theta  -\eta_t \nabla_{\theta}(\mathcal{L}_\textrm{mtrain}( {\theta}(w ))
\end{equation}
Based on such a fixed  classifier $\Phi_{\hat{\theta}(w)}$, the meta-objective is defined as the loss on the reference data: 
\begin{equation}
\label{eq:meta_obj}
\begin{split}
   \mathrm{MetaObjective}\left(\hat{\theta}(w )\right) & \triangleq \mathcal{L}_\textrm{ref}\left( \hat {\theta}(w) \right)\\
         & = \sum_{(x_i,r_i) \in \mathbb{D}_\textrm{ref}} \ell (\Phi_{\hat{\theta}(w)}(x_i),r_i).
\end{split}
\end{equation}
By applying gradient descent to minimize such meta-objective, we can optimize $w$ through second order derivatives. Thus, we get our new weight vector as:
\begin{equation}
\label{eq:max_w}
    w^\ast = \operatorname*{argmin}_{w: w\geq0}\mathrm{MetaObjective}\left(\hat{\theta}(w)\right) 
\end{equation}

With the learned instance weights $w^\ast$, we can update the classifier according to Eq. \ref{eq:train_model}. This phase is the actual \textbf{classifier training} phase:
\begin{equation}
\label{eq:train_model}
\theta^\prime = \theta - \eta_t \nabla_{\theta} \left(\sum_{(x_i,r_i) \in \mathbb{D}_\textrm{mtrain}}w^\ast_i \ell(\Phi_{\theta}(x_i),r_i)\right).
\end{equation}
The updated  $\theta^\prime$ 
are  used  as the parameters of $\Phi$ in the  next iteration. 

It is worth noting that $w$ is optimized to make the classifier perform better on $\mathbb{D}_\textrm{ref}$ as Eq.~\ref{eq:max_w} expresses. Further analysis~\cite{RenZYU18,shu2019meta} shows that the training instances whose gradient directions are similar to the gradient direction of $\mathbb{D}_\textrm{ref}$ will be assigned with high weights, otherwise low weights. With such bilevel optimization process, we could maximize the role of reference data for denoising.

\subsection{Enhancing Reference Data by Elite Instances}  
\label{sec:enhanced_sec}
The small amount of clean reference data could provide limited guidance when participating in the meta-objective, resulting in poor generalization capability. To overcome this issue, we augment the reference data by distilled elite  instances from the noisy data.
The elite instances are those that are attributed to the highest classification sores, thus are highly reliable. The elite instances are ``elected'' from the noisy data by  the classifier they trained. They are thus the most representative likely-true instances among the noisy data. The elite instances are intended to provide additional reliable supervision signals to the reference data, as we explained earlier.

\subsubsection{Enhancement strategy}
\label{ssec:s2v}
We propose to augment the reference set dynamically, i.e., 
we evaluate training data at each epoch and select top-scored instances to  expand the original reference set $\mathbb{D}_{\textrm{ref}}$ to $\mathbb{D}_{\textrm{ref}}^{\ast}$.  
Since the top-scored instances vary for each training epoch, we are able to leverage  diverse  elite instances to mitigate the ``mode collapse issue''.

We propose two strategies to compute the confidence score of labeling quality based on respectively the instance weighting ($sw$) and classification probability ($sp$). 
In the first strategy, we use the normalized instance weight determined with respect to the clean reference data (the same as in meta-learning) as the confidence score, defined as $sw_i = \frac{w_i}{\sum w_j} $. 
Since the instances assigned with high weights have gradient directions similar to the reference data~\cite{RenZYU18,shu2019meta}, $sw$ may help select instances consistent with the reference data. 

The second strategy uses the prediction probability by the classifier, defined by Eq.~\ref{eq:prob_score}.  
\begin{equation}
\label{eq:prob_score}
sp_i = \frac{P_{\Phi}(r_i|x_i)}{\sum P_{\Phi}(r_j|x_j)} 
\end{equation}
As the classifier is trained with a large amount of weighted noisy data, the instances whose labels are scored high by the classifier are more likely to be consistent with the common patterns emerged from boosted training data. 
The advantage of  using $sp$ is that its confidence score becomes more stable as the classifier training converges, while the confidence score by $sw$ is  learnt online and fluctuates across training epochs. We further examine this in Section~\ref{ssec:accumulative_score}.

With the confidence score $s_i$ (either $sw_i$ or $sp_i$), we rank and select instances from top to bottom. 
We select the top-scored instances for each type of relation, proportionally to the number of that relation in $\mathbb{D}_\textrm{ref}$. Note that we follow the same relation distribution as $\mathbb{D}_\textrm{ref}$ in order to avoid introducing  unexpected  bias from unbalanced relation distribution of $\mathbb{D}_\textrm{train}$. 

Formally, let $\mathbb{RE}$ be the set of target relation types excluding the not-the-target relation type\footnote{We do not select instances of not-the-target type, i.e. the negative training instances, since they are dominant in the $\mathbb{D}_\textrm{train}$, $\mathbb{D}_\textrm{ref}$. 
Selecting such instances has been empirically found useless in our preliminary experiments.}; $n_r$ be the number of selected instances for relation $r$, set as 
$k\#\{(x_i,r_i)|(x_i,r_i)\in\mathbb{D}_{\textrm{ref}},r_i=r\}$
\footnote{$\#\{.\}$ is the function of counting the number of a set.}, where $k$ is the expansion ratio. Then the expanded reference data is as follows:
\begin{equation}
    \label{eq:expand_reference}
    \mathbb{D}^{(t)}_\textrm{exp} = \bigcup \limits_{r \in \mathbb{RE}} \big \{ (x_i,r)| s_i 
    \geq  \tau( n_r), (x_i,r) \in  \mathbb{D}_\textrm{train} \big\},
\end{equation}
where $ \tau(n_r)$ is  the minimum confidence score of top $n_r$.

Finally, the augmented reference data is the union of the original one and the expanded one, i.e., $ \mathbb{D}^{\ast(t)}_\textrm{ref} = \mathbb{D}_\textrm{ref} \cup \mathbb{D}^{(t)}_\textrm{exp} $.
The meta-objective in Eq.~\ref{eq:meta_obj} is enhanced as:
\begin{equation}
    \label{eq:new_meta_obj}
    \mathrm{MetaObjective}\left(\hat{\theta}(w)\right) = \mathcal{L}_{\textrm{ref}}\left (\hat{\theta}(w)\right) +\beta \mathcal{L}_{\textrm{exp}}\left(\hat{\theta}(w)\right) ,
\end{equation}
where $\beta$ is a factor to control the influence of expanded reference data for meta-objective.

\subsubsection{Robust exploitation phase} 
\label{ssec:accumulative_score} 
The aforementioned strategy enhances instance reweighting, but may still suffer from wrong labels in the expanded reference set.
To further improve the quality of selected elite instances, we propose an additional exploitation strategy based on the assumption that the more times an instance has been selected as elite instance in different epochs, the more likely it is true. 
Since confidence scores among different training epochs are incomparable, we design an accumulate score from the ranking order to record the historical selection information. 
The accumulative score $sa_i$ for each training instance $(x_i,r_i)$  at training epoch $t$ is defined as follows: 
\begin{equation}
    \label{eq:accumu_score}
    sa_i^{(t)} = \left \{  
    \begin{aligned}
    & 0 , && \text{if}\ t=0\\
    & \gamma sa_i^{(t-1)} + sr_i^{(t)}, && \text{otherwise}
    \end{aligned} \right .
\end{equation}
where $\gamma$  is a parameter used to decay the influence of distant historical evaluations, 
and $sr_i^{(t)}$ is the ranking score 
at  epoch $t$:
\begin{equation}
\label{eq:rank_score}
 sr_i = \frac{1}{1+e^{\textrm{Idx}_i-n_r}}.
\end{equation}
where $\textrm{Idx}_i$ is its rank  according to $sw_i$ or $sp_i$ from top to bottom. 

We apply the exploitation strategy only after a certain number of training epochs (in our experiments, half of the total training epochs) in order to allow for more diversified selection at the beginning (\emph{aka} exploration). 

\subsection{Algorithm}
We design an iterative algorithm for model training, as shown in Algorithm~\ref{alg:overall_training}.
At the training epoch $t$, we compute the current confidence score $sp_i$~(line 4) (which could also be $sw_i$) and update it to the accumulative score $sa_i$~(line 5), which is used in exploitation phase.
Then we select top-scored instances according to $s_i$ to construct expanded reference set~(line 8).
\begin{algorithm}[thb]
	\centering
	\caption{Our approach for robust classifier training from noisy data.}
	\label{alg:overall_training}
	\begin{algorithmic}[1]
		\REQUIRE
		Noisy training dataset $\mathbb{D}_\textrm{train}$; \\
		\hspace{7mm} Clean reference dataset $\mathbb{D}_\textrm{ref}$; \\
		\hspace{7mm}  Classifier $\Phi$ parameterized by $\theta$\\
		\ENSURE 
		$\theta ^{(L)}$
		\STATE
        Pre-train $\Phi$\ (with parameters $\theta ^{(0)}$);  \\
		\FOR{epoch $t=1$ to $L$}
		    \FOR{\textbf{each} $(x_i,r_i)$ in $\mathbb{D}_\textrm{train}$}
		    \STATE Compute confidence score $sp_i$ with $\Phi(\theta^{(t)})$ as Eq.~\ref{eq:prob_score}.
		    \STATE Accumulate ranking score $sa_i^{(t)}$ 
		    as Eq.~\ref{eq:accumu_score}
		    \STATE Set final score $s_i$ as $sa_i^{(t)}$ if in exploitation phase otherwise $sp_i$~(Section.~\ref{ssec:accumulative_score})
		    
		    \ENDFOR
		    \STATE Select top-scored instances based on $s_i$ to obtain expanded reference set
		    $\mathbb{D}_\textrm{exp}^{(t)}$~(Section~\ref{ssec:s2v})
		    \FOR {each mini batch in $\mathbb{D}_\textrm{train}$}
		        \STATE Calculate meta-objective with the loss on  $\mathbb{D}_\textrm{ref}$ and $\mathbb{D}_\textrm{exp}^{(t)}$ as Eq.~\ref{eq:new_meta_obj}. 
		        \STATE Optimize $w^\ast$ based on the meta-objective as Eq.~\ref{eq:max_w}.  
		        \STATE Use $w^\ast$ to update classifier $\Phi(\theta)$ as Eq.~\ref{eq:train_model} 
		    \ENDFOR
		\ENDFOR
	\end{algorithmic}
\end{algorithm}

Then the augmented reference data is applied to the online reweighting algorithm, which updates $w^\ast$ and $\theta$ using mini-batches of training data~(lines 11 - 12), similar to the existing meta-reweighting approach~\cite{RenZYU18,shu2019meta}. 
\section{Experimental Setup}

\subsection{Datasets}
\label{sec:data_and_settings}
We evaluate our approach on two widely-used DS datasets for relation extraction/classification: Wiki-KBP and NYT. To evaluate our models precisely, we construct a new  version of Wiki-KBP and adopt a manually labeled test set for NYT (see explanations below).
The statistics about the datasets  are presented in Table~\ref{tab:datasets}. 
 
\begin{table}[htb]
  \caption{Statistics of the datasets used in our experiments.}
  \label{tab:datasets}
  \begin{center}
  \resizebox{0.95\columnwidth}{!}{
  \begin{tabular}{llcrr}
    \toprule
     \multicolumn{2}{c}{Datasets}  &  \#Relation types$^{\mathrm{\ast}}$ & \#Instances & \#Positive instances\\
    \midrule
    \multirow{2}{*}{Wiki-KBP}&Training  & 6 &151,091 & 38,922\\
    & Test & 6 & 4,168 &1,075 \\
    \midrule
    \multirow{2}{*}{NYT} & Training & 53 & 522,611 &136,947\\
    & Test & 17 & 2,040 & 684\\
    
  \bottomrule
  \multicolumn{5}{l}{$^{\mathrm{\ast}}$Including the Not-Target-Type for negative instances.}
 \end{tabular}
}
\end{center}
\end{table}
\emph{Wiki-KBP} 
The training dataset was constructed by aligning Freebase facts with English Wikipedia corpus~\cite{ling2012fine}.   
The commonly-used test set was manually annotated from the 2013 KBP corpus~\cite{ellis2012linguistic}. 
However, the relation types in the test set do not always match with those in the training set - some test relation types  have no or only one training instance. We thus removed these relation types. 

The resulting test set contains a very small number of positive instances~(165) and their proportion in the test set is only about 0.08. To have more labeled sentences of target relations for evaluating classifiers, we increase the number of positive instances in the test set by merging another manually annotated dataset - TACRED~\cite{zhang2017position}, whose sentences are also derived from the KBP corpus. 

Specifically, we removed the relation with only one instance in the training set~(per:countries\_of\_residence), and obtained the modified training set with 5 actual relations - per:country\_of\_death, per:country\_of\_birth, per:children, per:parents, per:religion, and a special relation not-target-type~(None) for negative instances. Then, we merged the positive instances of these 5 types from the original test set~\footnote{https://github.com/shanzhenren/CoType} and the TACRED dataset~\footnote{https://nlp.stanford.edu/projects/tacred/}.
To make sure that our test set has a similar proportion of positive instances as the training set, we randomly sampled 3093 negative instances~(``no\_relation'') from the TACRED dataset. The resulting training  and test sets are as shown in Table~\ref{tab:datasets}.

\emph{NYT} This dataset was generated by aligning news corpus from New York Times (NYT)\footnote{http://iesl.cs.umass.edu/riedel/ecml/} and relation facts in Freebase
~\cite{riedel2010modeling}. Most previous work used held-out evaluation, where the training set and test set were heuristically annotated based on disjoint sets of the freebase facts. 
However, the DS-generated labels could be wrong and the two sets are noisy. 
To evaluate the relation classification at sentence level, we used the original training set and a manually annotated test set provided by \citeauthor{jiang2018revisiting} \cite{jiang2018revisiting}.

The \textbf{clean reference data} is constructed as follows: For Wiki-KBP, to better evaluate our models, we  split the original test dataset into a validation set and a test set. We fixed the test set and randomly sampled \textbf{100 instances} from the clean validation set as the reference data. 
For NYT, since the manually labeled data is very small, we sampled 10\% of the training data as validation set and left the rest as the actual training set. 
To keep the original setting in which no entity pair overlaps between the training and test sets, we randomly sampled one fifth entity pairs from the test dataset and extracted the corresponding sentences~(\textbf{about 400 instances}) as the reference data. The remaining test data was used as actual test set.

Note that we set the minimal size of reference data by making sure that there is at least one instance for each relation in the reference set. 
We randomly sampled five sets of reference data and fixed them to conduct our experiments. All the reported average results for baselines and analysis experiments are based on the same reference sets. 

\subsection{Compared models}
\subsubsection{Our model}
Based on the model-agnostic meta-learning algorithm, our approach can be applied to any neural network architectures for DS relation classification. 

In this work, we adopt the Piecewise Convolutional Neural Network (PCNN)~\cite{zeng2015distant} as base model to compare our approach with other denoising approaches. 
This model has been widely used for relation classification with DS~\cite{zeng2015distant,lin2016neural,ji2017distant,wu2017adversarial} and been proven to perform better than CNN models~\cite{jiang2018revisiting}.
The PCNN model contains an input layer which concatenates the word embedding and the position embedding, a convolutional layer, a piecewise max pooling layer, and a softmax layer. More details can be found in~\cite{zeng2015distant}.

Based on the general meta-learning mechanism described in section~\ref{sec:metalearning}, we build our approach on top of learning to reweight examples~(L2RW)~\cite{RenZYU18} algorithm. It regards instances weights as a meta-parameter vector implicitly learned and does not require extra hyper-parameter tuning. Instance weights can also be learned by a parameterized module within the meta-learning framework, such as a multilayer perceptron network~(Meta-Weight-Net)~\cite{shu2019meta}. We will not examine it in this paper.

We measure the effectiveness of two strategies for elite instance selection, i.e. $sw$~(by online-learnt instance weights) and $sp$~(by prediction probability from the classifier). Then we use $sp$ as the default to select elite instances for detailed analysis in section~\ref{ssec:ExpAnalysis}.

\subsubsection{Baselines}
We compare our approach with previous representative and state-of-the-art instance selection approaches in DS relation classification, including:
\begin{itemize}[leftmargin=*]

    \item \textbf{PCNN+ATT}~\cite{lin2016neural} is a classical bag-level instance weighting approach. It  assigns attention weights to each instance within a bag according to their relevance to the bag label, 
    and thus down-weights the relatively noisy instances.
  \item \textbf{PCNN+RL}~\cite{WangXQ18robust} adopts reinforcement learning to generate the false-positive indicator to recognize false positives, and then redistributes them to the negative set to obtain a new cleaned dataset. 
  \item \textbf{PCNN+DSGAN}~\cite{WangXQ18dsgan} adopts adversarial learning to train a generator to recognize true positive instances, and then redistributes the remaining false positives to the negative set to obtain a new cleaned dataset. 
  \item \textbf{ARNOR}~\cite{jia2019arnor} is previous state-of-the-art model for DS relation classification. It starts the model training with reliable instances selected by a set of frequent relation patterns, and then adds  patterns by bootstrapping.
\end{itemize}

We apply the reference data to enhance existing models as follows: 1) for RL~\cite{WangXQ18robust} and DSGAN~\cite{WangXQ18dsgan}, we compute reward on reference data with the same reward function and average it with previous rewards to obtain final feedback for instance selection; 2) for ARNOR~\cite{jia2019arnor}, we add the relation patterns extracted from the reference set to their pattern collections. 

We compare with the following baseline models that use extra clean data:
\begin{itemize}[leftmargin=*]
  \item \textbf{PCNN+L2RW}~\cite{RenZYU18} is the learning to reweight samples algorithm that resorts to a clean unbiased dataset to tackle the biased training set problem. 
  \item \textbf{PCNN+BA}~\cite{ye2019looking} is a bias adjustment approach that uses the extra clean data to help relieve the label distribution shift from DS training data and manually annotated test data. We use their best-performing BA-Fix model.
  
\end{itemize}

\subsection{Implementation Details}

We use the same word embedings for baselines, i.e., the pre-trained GloVe~\cite{pennington2014glove} embedding\footnote{http://nlp.stanford.edu/data/glove.840B.300d.zip} for Wiki-KBP and the 50-dimensional word embedding file\footnote{https://github.com/thunlp/NRE} released by \citeauthor{lin2016neural}~\cite{lin2016neural} for NYT. The corresponding position embeddings are set as 30 dimensions for Wiki-KBP and 5 dimensions for NYT. The number of convolution filter for PCNN model is 230, and the filter window size is 3.

For our enhanced reweighting strategy, we select $k$ times elite instances compared to the original reference data. 
$k$ falls between 2 and 3, depending on the dataset.
Parameter $\beta$ is generally set as 1, and adjusted as 0.1 for $sw$ during the exploitation phase. 
$\gamma$ is  set as 0.97. 
For model training, we  adopt the SGD optimizer and set the maximum training epoch as 25. 
The batch size for training set is 160 and we use all the reference data for every reweighting step due to its small size.  
To speed up the training process, we linearly warm up the learning rate to a maximum value of 0.1 within 2 epochs, and linearly decrease it to a minimum value of 0.001 used for the last 5 epochs.

All the baselines were implemented with the source codes released by their authors except for ARNOR - ARNOR is not open-sourced, so we reimplemented it based on an open implementation\footnote{https://github.com/HeYilong0316/ARNOR}, which achieves comparable performance on the same dataset as reported in~\cite{jia2019arnor}. Since ARNOR starts by selecting reliable instances based on several frequent relation patterns, when  training instances are spread over many
patterns~(e.g. in Wiki-KBP), it fails to select enough training data.
Therefore, we increased the number of patterns when applied it to our new dataset to increase its performance.

\section{Evaluation Results}
We use the same metrics as previous work~\cite{zhang2017position,jia2019arnor}: micro-averaged Precision~(Pre.), Recall~(Rec.) and F1-score~(F1).

\subsection{Overall Results}
\label{ssec:overallResults}
\begin{savenotes}
\begin{table*}[tp]
  \caption{Test set performance comparison of different models on the Wiki-KBP and NYT  datasets. The five-time average and standard deviation of test results are reported as percentage, and the best~(\textbf{bold}) and second best~(\underline{\textit{italic}}) F1 scores are highlighted below.}
  \label{tab:test_results}
  \begin{center}
  \begin{tabular}{l|ccc|ccc}
    \hline
     \multirow{2}{*}{Model}  & \multicolumn{3}{c|}{Wiki-KBP}&\multicolumn{3}{c}{NYT}\\ \cline{2-7}
      & Prec. & Rec. & F1 & Prec. & Rec. & F1 \\ \cline{1-7}

    PCNN$^{\triangle}$~\cite{zeng2015distant} 
    &$55.39 \pm 3.75$ & $34.37\pm 4.14$ &$42.16 \pm 1.97$
    & $46.93\pm 1.68$ & $57.79\pm 5.01$ & $51.66\pm1.61$  \\
  
    PCNN~\cite{zeng2015distant} 
    &$56.12 \pm 3.33$ & $33.38\pm 2.17$ &$41.75 \pm 0.98$
    &$47.88\pm 1.81$ & $57.38\pm 3.75$ & $52.12\pm1.41$ \\
    PCNN+ATT~\cite{lin2016neural} 
    & $72.65\pm 1.99$ & $29.24 \pm 1.31$ & $41.69 \pm 1.55$
    & $59.99\pm1.86$ & $49.79\pm2.29$ & $54.36\pm0.79$ \\ 
    PCNN+RL~\cite{WangXQ18robust} 
    &$57.64 \pm 2.41$ & $38.79 \pm 2.22$ & $46.32\pm 1.65$
     & $48.72\pm1.72$ &	$48.93\pm2.11$ &	$48.78\pm0.46$  \\
   
    PCNN+DSGAN~\cite{WangXQ18dsgan}
    &$59.86 \pm 5.65$ & $38.54 \pm 2.97$ & $46.65 \pm 1.19$
     &$47.55\pm1.15$ &	$51.52\pm1.30$ &	$ 49.44\pm0.75$ \\
    ARNOR~\cite{jia2019arnor}
    &$54.83\pm2.40$&$34.59\pm 2.20$&$42.35\pm 1.32$ 
    &$68.39\pm1.37$ & $ 48.42 \pm 2.66 $ &$ 56.67\pm 2.06 $ \\
    \hline
     PCNN+BA~\cite{ye2019looking} 
     &$58.86\pm2.92$ &	$38.71\pm2.60$ &$46.59\pm2.73$ 
     & $45.33\pm0.86$ &$58.06\pm 3.24$ &$50.70 \pm 1.21$ \\ 
 
    PCNN+L2RW~\cite{RenZYU18} 
    
   &$66.22\pm 3.36$&$43.80\pm 4.72 $&$52.56\pm 3.45$ 
    &$ 66.23\pm4.17$  &	$ 53.05\pm 3.40$  &	$ 58.56\pm 2.12$ \\ 
   
   OURS\_sw 
   &$60.15\pm3.58$&$51.43\pm5.05$&$\underline{\mathit{54.56^{\ast}\pm3.94}}$
    &$ 63.65 \pm 4.63$ &$57.79\pm4.54$&$\underline{\mathit{60.38^{\ast}\pm2.54}}$\\
    OURS\_sp 
    &$ 67.24 \pm 3.10$ & $46.65 \pm 4.13$ &$\mathbf{54.98^{\ast}\pm3.00}$ 
    &$ 67.05 \pm 4.55$ &$56.59\pm2.25$&$\mathbf{61.26^{\ast}\pm1.98}$\\  
 
    \hline
    \multicolumn{7}{l}{$^{\triangle}$ represents the model trained without using clean  data. }\\
    \multicolumn{7}{l}{ $^{\ast}$ indicates statistically significant improvements
    over the L2RW~(i.e. Wilcoxon signed-rank test with $p < 0.05$)\footnotemark.
    }
    
\end{tabular}
\end{center}
\end{table*}
\end{savenotes}
\footnotetext{
We use the Wilcoxon signed-rank test to measure the paired results of the L2RW and OURS based on the same set of clean reference data, and the t-test results are computed with the scipy package.
}
From the results shown in Table~\ref{tab:test_results}, the following observations can be made:

\begin{itemize}[leftmargin=*]

  \item Our approach outperforms all the baselines on F1 score and improves the F1 score of base PCNN model~(line 2) by over 12\% on Wiki-KBP and 8\% on NYT. This  demonstrates the higher effectiveness of our approach.
  
  \item L2RW and our approach outperform other instance selection approaches. This shows that meta-learning based approaches are able to make better use of the clean reference data than simply adding it to the training data~(PCNN in line 2), using its relation patterns to extract training data~(ARNOR), and  computing reward for instance selection under existing RL or DSGAN framework. We also empirically found that fine-tuning the PCNN model with clean reference data does not improve its performance and may even decrease it when the size of  clean reference data is too small. 
  \item 
  Both strategies of our approach improve L2RW by a statistically significant margin.
  This shows that our enhanced strategy for instance reweighting is beneficial. We observe that L2RW can substantially improve the precision  of PCNN model on two datasets, but decrease the recall on NYT.
  This observation confirms our intuition that the small size of clean reference data may lead to mode collapse. 
  In contrast, with our enhanced strategy, the recall score and F1 score are consistently increased.
  This indicates that the expanded reference data could enhance the reweighting process to select more reliable training data with new patterns, thus improving the generalization capability.
    \item The bias adjustment approach~(PCNN+BA model) does not show stable improvements on two datasets compared to the base model. The F1 score of PCNN+BA is slightly decreased on NYT.
    We explain this by the fact that the reference set and test set in NYT are divided based on entity pairs, and thus the reference set does not provide exactly matched label distribution information.
    This shows that the small set of clean reference  data could provide little information about label distribution, but our approach using it for denoising makes a big difference.
\end{itemize}

\subsection{Analysis and Discussion}
\label{ssec:ExpAnalysis}
\subsubsection{Understanding the impact of reference data}
Our approach reweights the noisy training data under the guidance of augmented reference data during the classifier training. To understand the impact of reference data, 
we present the performance of classifiers that are trained under three settings: with original noisy training data~(PCNN), with meta-reweighting algorithm guided by clean reference data~(L2RW), with meta-reweighting algorithm guided by augmented reference data~(OURS). 

\begin{figure}[bhpt]
\centering
\includegraphics[width=8cm]{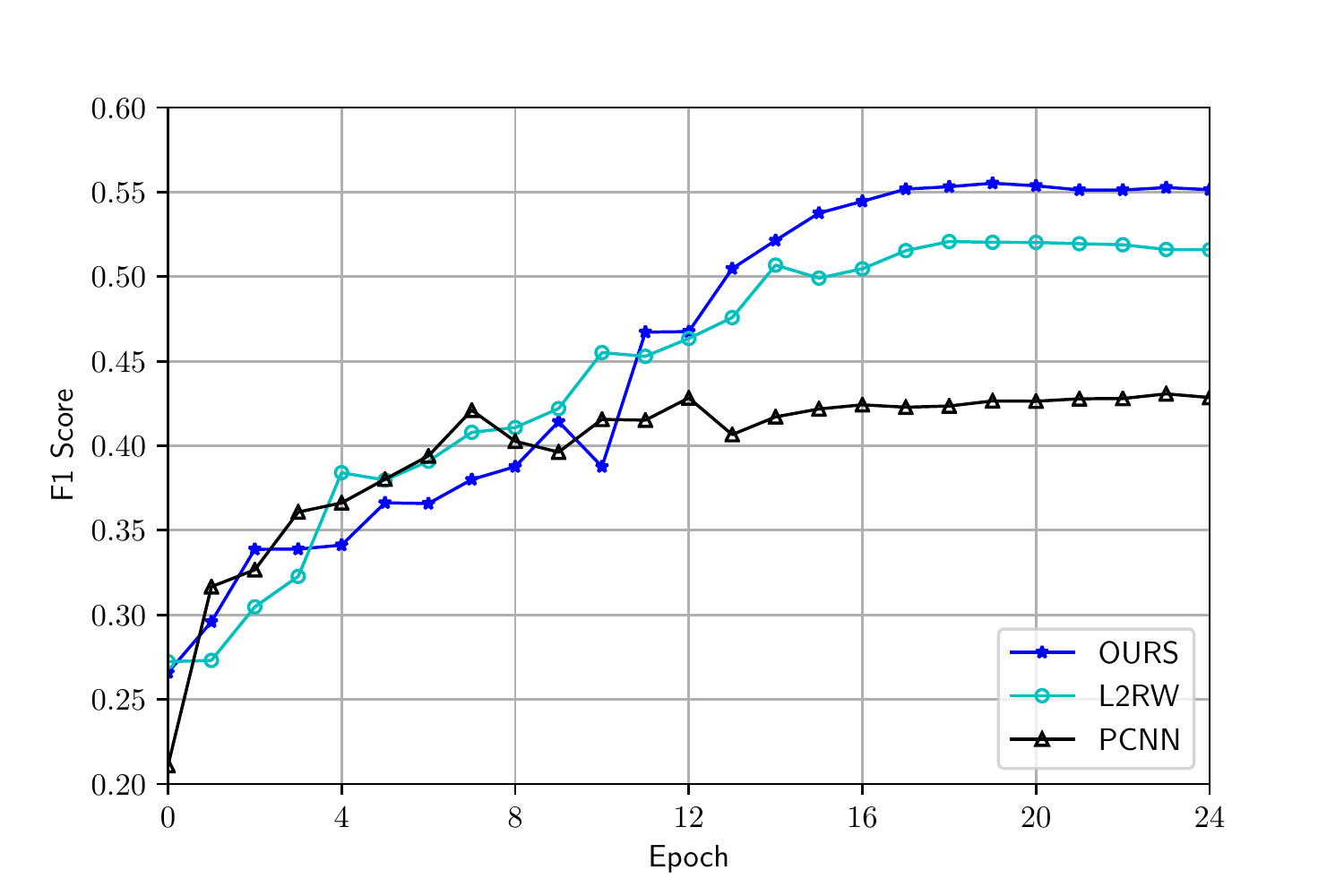}
\caption{Test set F1 score of three models on Wiki-KBP at each epoch. The lines are drawn with average F1 scores based on five reference sets.}
\label{fig:training_performance}
\end{figure}

As shown in Fig~\ref{fig:training_performance}, after several training epochs, PCNN model quickly stagnates, while our approach and L2RW continue to improve the performance with a large margin. This shows the impact of the meta-reweighting for learning under DS.
During the exploitation phase~(after epoch 12), our enhanced strategy exploits $sa$ to distill elite instances and uses them to guide the reweighting process. We see steadily superior performance over L2RW
. This is a demonstration of the usefulness of elite instances. 

During the exploration phase~(before epoch 12), our augmenting strategy selects elite instances  according to the current confidence score. 
Fig.~\ref{fig:training_performance} shows that it may perform poorer than L2RW sometimes. This may be due to the fact that the selected elite instances may contain noise. 
As the training epoch increases, the well-trained classifier and  $sa$ could help distill more reliable elite instances, thus enhancing the reweighting. At the end, our final model after all the training epochs is better than the others.

\subsubsection{Impact of exploitation strategy}
\label{sssec:analyze4evluation_criteria}
 To reduce possible false instances in elite data when using the current confidence score~(i.e. $sw$ or $sp$) for selection, we propose to use accumulative score~($sa$) during the exploitation phase~(after epoch 12). To verify the effectiveness of the exploitation strategy, we test our approach with and without using the accumulative score~(with $sa$ vs. w/o $sa$) and compare them with the algorithm without enhancement strategy~(L2RW). 
 We apply meta-reweighting algorithms to train the PCNN model 
 based on one set of reference data and present their smoothed convergence curves of 5-step moving averages on both the validation and test sets of Wiki-KBP. 
 
 \begin{figure}[hbt]
     \centering
     \includegraphics[width=\linewidth]{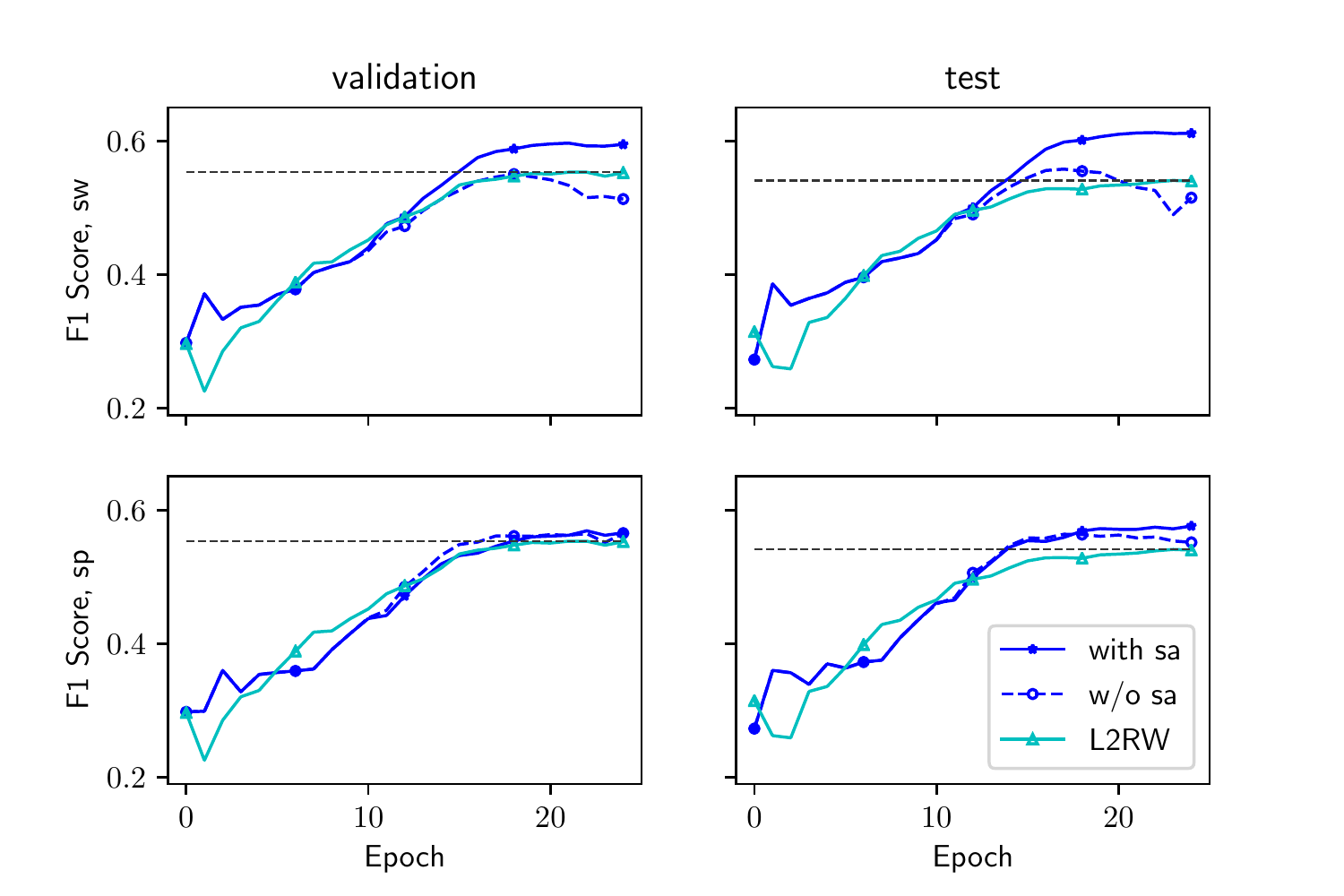}
     \caption{Smoothed convergence curves of F1 scores for models using $sa$ or not on both validation set and test set of Wiki-KBP. Results for two evaluation criteria $sw$ and $sp$ are presented respectively.}
     \label{fig:eva_confscore}
 \end{figure}
 
As shown in Fig.~\ref{fig:eva_confscore}, $sa$ is beneficial for both evaluation criteria~($sw$ or $sp$) and is especially important for $sw$. When $sa$ is not used, the performance of models using $sw$ decreases at the end of training process and may be even poorer than L2RW sometimes. This shows that noise elite instances may be selected into the expanded reference set. When using $sa$, more reliable elite instances are selected. The improvements are more steady.
Note that when using other reference sets the improvement margin over L2RW may vary, but similar impact of the exploitation strategy is observed.
The above observation shows that the exploitation strategy that takes into account the historical evaluations is beneficial, especially for the scenario where evaluation results fluctuate across epochs.

\subsubsection{Size of the clean reference data}
\label{sssec:varyRefSizeCompare}
\begin{figure}[bhpt]
\centering
\includegraphics[width=0.85\linewidth]{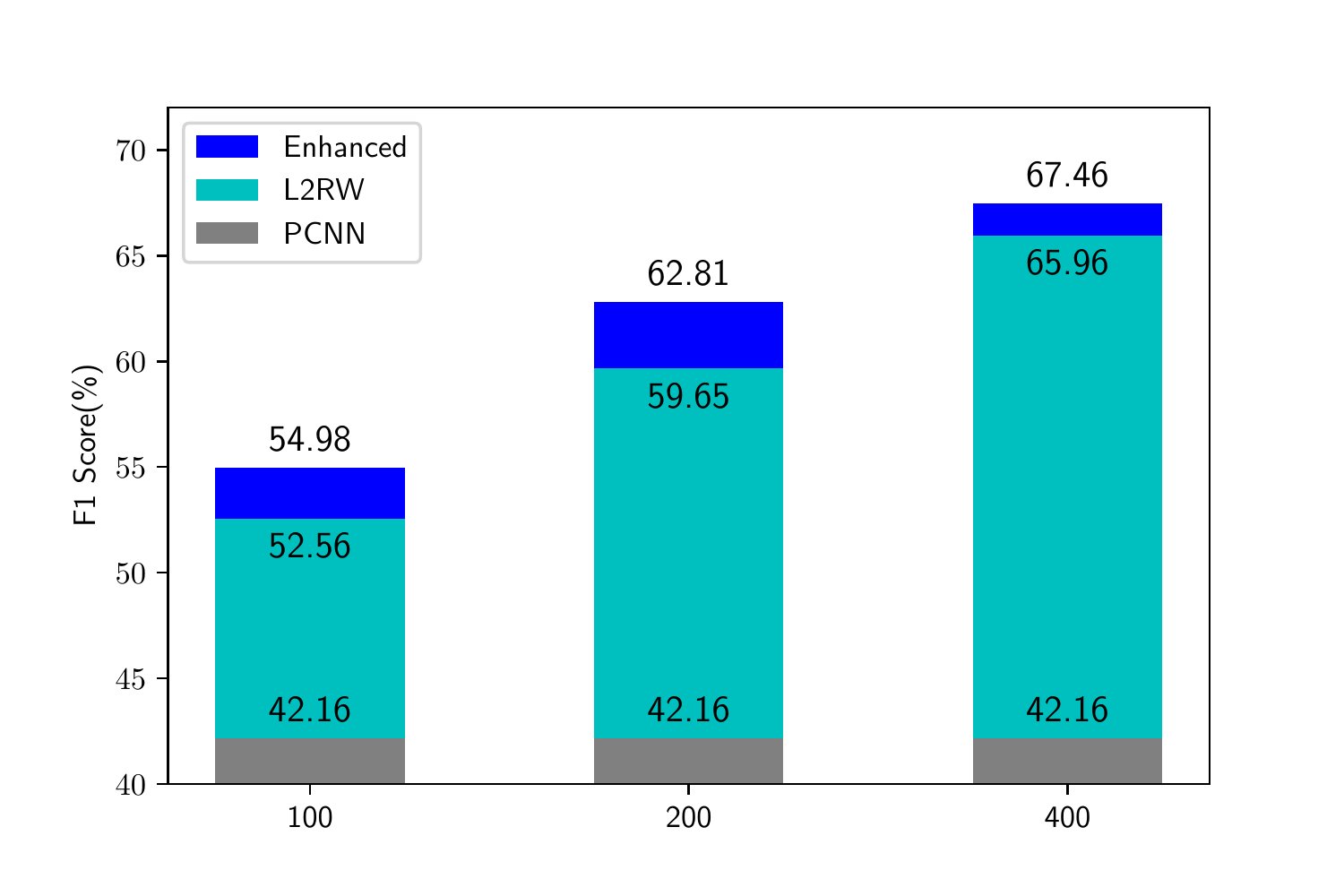}
\caption{Test set F1 score improvements compared with PCNN model when different sizes of clean reference data are used by our approach in Wiki-KBP dataset. 
}
\label{fig:vary_clean_Data}
\end{figure}

To see the impact of the size of initial clean reference data on meta-reweighting algorithm, we evaluate 
the performance of models when the size of clean reference data is increased from the initial 100. Following the same setting, we randomly sample the required reference set five times from the validation set of Wiki-KBP, and present the average results with L2RW and enhanced strategy in Fig.~\ref{fig:vary_clean_Data}. 
We can see that 
when the number of clean reference data increases, both our model and L2RW can further improve the performance of the classifier. When 400 clean reference samples are available, our approach achieves 67.46\% F1 score, which improves the basic PCNN model by more than 25\%. However, we also see that the gap with L2RW is reduced. This suggests that 
our enhancement with elite data is more useful when the size of the clean reference data is relatively small. We explain this by the fact that when there is sufficient clean reference data, it becomes less critical to further extend it because it already has a quite good coverage of different relation patterns. So, our enhancement is the most beneficial with a small amount of clean reference data - the situation we target in this paper.

\subsubsection{Case study}

To better understand the impact of the meta-reweighting algorithm and our expanded reference instances, we analyze the case of one particular relation. 
The case shown in Table~\ref{tab:augment_sents} is an  example extracted from our data for relation `/location/administrative\_division/country' in NYT dataset, and Table~\ref{tab:case_study} presents examples about sentence weighting and relation classification. 
To make our analysis easier, we only consider using  similar words between two entities to match sentences, which is also regarded as the relation pattern in ARNOR~\cite{jia2019arnor}.

\begin{table}[tbh!]
  \begin{center}
  \resizebox{0.98\columnwidth}{!}{
  \begin{tabular}{lll} 
    \toprule
    \multicolumn{3}{l}{Relation: /location/administrative\_division/country}\\
     \midrule
     Dataset&ID &Sentences\\
     \midrule
    \multirow{4}{*}{Clean reference} & 1 &... a company in \textbf{Auroville}, \textbf{India} \\
  & 2 &... on the school in\textbf{ Beslan}, in southern \textbf{Russia } \\
  & 3 &... ``cities on the rise'': \textbf{Riga}, \textbf{Latvia}\\

    \midrule
    \multirow{2}{*}{Expanded reference} 
    & 4 &... for nearby \textbf{Aleppo} \underline{in} \textbf{Syria} \\
   & 5 &... family's zoo in \textbf{Queensland}, \underline{in northern} \textbf{Australia}\\
    \bottomrule
\end{tabular}
 }
 \caption{Examples showing that our elite instances enrich patterns of clean reference data.}
 \label{tab:augment_sents}
\end{center}
\vspace*{-0.5cm}
\end{table}

\begin{table}[tbh!]
  \begin{center}
  \resizebox{0.98\columnwidth}{!}{
  \begin{tabular}{llccc} 
    \toprule
    \multicolumn{4}{l}{Relation: /location/administrative\_division/country}\\
     \midrule
     ID& Sentences  & ARNOR & L2RW & OURS\\
     \midrule
      
    6 &  ... tour in \textbf{Chechnya} crashed \underline{in southern} \textbf{Russia} & 0 &0.002 &0.003 \\
    7 &  ... 
    in outback \textbf{New\_south\_wales} \underline{in} \textbf{Australia} & 1 &0.0 &0.004  \\
    8 &  ...  \textbf{Haryana} and Uttar Pradesh \underline{in northern} \textbf{India} ... 
    &0&0.0&0.003 \\

    \midrule 
   9&... winter olympics in \textbf{Albertville} \underline{,} \textbf{France}. &$\times$&$\surd$&$\surd$\\
   10 &...   a Philistine  seaport at \textbf{Ashkelon} \underline{in} \textbf{Israel}. &$\surd$&$\times$&$\surd$ \\
    11 &...  here in rural \textbf{Bihar} state \underline{in northern} \textbf{India} ... &$\times$&$\times$&$\surd$\\
    \bottomrule
\end{tabular}
 }
 \caption{Examples about sentence weighting~(the upper part) and relation classification~(the lower part). We underline similar relation expressions with the reference samples.
 }
 \label{tab:case_study}
\end{center}
\vspace*{-0.5cm}
\end{table}

\textbf{Effectiveness of meta-reweighting.} The relation patterns of clean reference data are all used by ARNOR to select training data with the same relation patterns, such as ``entity1 , in southern entity2'' in sentence 2. However, it fails to select some useful training sentences, such as sentences 6 and 8, due to their different patterns. In contrast, the reweighting algorithms~(L2RW and OURS) can assign a positive weight to sentence 6 which is similar to sentence 2. 
This confirms that the meta-reweighting algorithms could make better use of the limited supervision signal~(the clean reference data) to boost relevant training data than  using the shallow  patterns to select training data~(ARNOR).

\textbf{Effectiveness of expanded reference data.}
Table~\ref{tab:augment_sents} shows that the distilled elite instances from noisy training data could provide not only similar relation expression (sentence 5 which is similar to sentence 2), but also new expression patterns such as ``entity1 in entity2'' in sentence 4. This shows that the expanded reference has the capability of covering more relation expression cases. 

Once our distilled elite instances (e.g. sentence 4 and sentence 5) are added into the reference set, sentences 6-8 in Table~\ref{tab:case_study} are weighted positively (i.e. selected as training instances for the classifier) because they bear some similarity with the expanded reference data. As a result, the test sentences 9-11 are all correctly classified.
This examples show the underlying reason why our enhancement of reference data can improve the classification effectiveness.

\section{Related Work}
\subsection{Distantly Supervised Relation Classification}
Relation Classification is a fundamental task in natural language processing. Neural network based models have achieved state-of-the-art performance on this task~\cite{zeng2014relation,zhou2016attention,wang2016relation,zhang2017position}. However, 
training effective neural classifiers requires a large amount of labeled data, which is usually hard to obtain. 
Distant supervision provides a way to create massive weakly labeled data for relation classification but the inevitable wrong labels harass reliable training~\cite{mintz2009distant,riedel2010modeling}.

Most existing studies train relation classifier in DS by applying multi-instance learning~(MIL) to reduce the impact of wrong labels~\cite{riedel2010modeling,hoffmann2011knowledge,surdeanu2012multi,zeng2014relation,lin2016neural,ji2017distant,liu2017soft}, which relaxes the relation label of each instance to a \textit{bag} of sentences containing the same entity pair.
Assuming at least one sentence within the bag expresses the target relation, MIL-based approaches train and test the relation classification at bag level. They generally face two limitations. (1) They still suffer from noises when all instances within a bag are false~\cite{WangXQ18dsgan,WangXQ18robust,li2019self}. 
Some recent studies mitigate this issue by incorporating complex attention modules among multi-bags ~\cite{li2019self,ye2019distant}. (2) These MIL based approaches~\cite{qu2019fine,li2019gan} are designed and tested for relation extraction, i.e. to extract all possible relations between a pair of entities, and they are not suitable for sentence-level relation prediction~\cite{feng2018reinforcement,jia2019arnor}.
In our work, we focus on relation classification at sentence level. Nevertheless, we compared with one representative MIL based approach~\cite{lin2016neural} in the experiments.

Alternatively, some recent studies evaluate and select training instances individually without relying on the at-least-one assumption~\cite{feng2018reinforcement,WangXQ18robust,WangXQ18dsgan,han2018denoising,zeng2018large,jia2019arnor,yang2019exploiting}, and our work belongs to this family. Previous approaches in this line rely only on the noisy training data to learn instance selection and may suffer from noisy labeling problem. In our work,  we introduce a small amount of clean data to guide the  instance selection.  
In addition, previous studies either rely on specific neural networks~(e.g. LSTM)~\cite{jia2019arnor}, or manually crafted reward functions~\cite{feng2018reinforcement,WangXQ18robust,WangXQ18dsgan}, while our approach adopts a model-agnostic meta-learning algorithm without manually specifying any specific form of reward functions or extra models, and thus is more widely applicable in practice. 

There are other studies that combine direct supervision and DS. \citeauthor{pershina2014infusion}~\cite{pershina2014infusion} introduced a small amount of labeled data by leveraging manually selected features from it, while we do not require manually selected features.
\citeauthor{beltagy2019combining}~\cite{beltagy2019combining} rely on large additional supervised datasets~\cite{beltagy2019combining} to help identify whether a sentence expresses a relation, while we only use a small set of labeled data as guidance. 

A recent work~\cite{ye2019looking} also use manually labeled data for relation classification in DS. They use it to adjust the DS-trained models for relieving the shifted label distribution problem. Our goal is different: we use it to select reliable instances to train effective neural models.

\subsection{Gradient-based Meta-learning}
Meta-learning~\cite{thrun1998learning}, also know as learning to learn, has made great progress with recent advances of gradient-based meta-learning~\cite{finn2017model}. 
It has wide applications, such as model parameter initialization~\cite{finn2017model,nichol2018first}, learning unsupervised update rules~\cite{metz2018meta}, learning sample weighting schema~\cite{RenZYU18,shu2019meta} and so on.

Some recent studies apply such meta-learning algorithm to learn model parameters for specific relation classification tasks. They focus on supervised relation classiﬁcation with limited supervision~\cite{obamuyide2019model} or lifelong relation extraction~\cite{obamuyide2019meta}, while we study DS-based relation classification.

Two closely related approaches~\cite{RenZYU18,shu2019meta} adopt the same gradient-based meta-learning mechanism as ours to tackle the training set bias issues. 
They use an ideal validation set which is clean, unbiased and can be scaled to the desired amount to reduce the bias between the training set and test set for image classification tasks. However, this kind of validation set is difficult to obtain in reality. In the DS relation classification task, the available clean reference data is much less than the automatically generated DS training data~(the proportion is less than 0.08\%), which might lead the model to mode collapse issues. In contrast, by distilling highly confident instances from noisy data to augment the reference dataset, our approach can leverage more reliable data during training and this effectively improves the performance, as shown in our experiments (Fig~\ref{fig:training_performance}).

\section{Conclusion}
In this paper, we proposed a meta-learning based approach for distantly supervised relation classification under the guidance of a small set of clean reference data. Our approach learns to reweight instances by minimizing the loss on a dynamically augmented reference set. This process  is able to leverage reliable training data and enhances the generalization capability. 
Experimental results on two distantly supervised datasets show that our approach outperforms previous state-of-the-art noise reduction approaches as well as the  meta-reweighting baseline.

Since the reference data in our approach plays as a key role for selecting reliable training instances, in the future, we plan to investigate active learning for annotating some representative reference samples and extend our approach to large scale relation extraction. Other criteria, such as diversity, could also be incorporated in instance selection.

\begin{acks}
The project is supported by the National Key R\&D Program of China (Contract No. 2018YFB21011100) and the National Natural Science Foundation of China~(61932001). We would like to thank Jiyang Zhang for his help in the implementation.
\end{acks}

\bibliographystyle{ACM-Reference-Format}
\bibliography{reference}
\end{document}